\documentclass[10pt,twocolumn,letterpaper, dvipsnames, fleqn]{article}

\usepackage{iccv}
\usepackage{times}
\usepackage{epsfig}
\usepackage{graphicx}
\usepackage{amsmath}
\usepackage{amssymb}
\usepackage{graphicx}
\usepackage{booktabs}
\usepackage[dvipsnames]{xcolor}
\usepackage{color, colortbl}
\usepackage{graphicx}
\usepackage{amsmath}
\usepackage{amssymb}
\usepackage{booktabs}
\usepackage{nicematrix}
\usepackage{makecell}
\usepackage{blindtext}
\usepackage{multirow}
\usepackage{bm}
\usepackage{bbm} % \mathbbm
\usepackage{comment}
\usepackage{algorithm}
\usepackage{algorithmic}
\usepackage{subcaption} %%%% this one is for the subtables 
\usepackage{array} % for extended column definitions, e.g. >{\em}c
\makeatletter
\renewcommand\paragraph{
  \@startsection{paragraph} % name
  {4} % level
  {\z@} % indent
  {.5em \@plus1ex \@minus.2ex} % beforeskip
  {-1.5em} % afterskip
  {\normalfont\normalsize\bfseries} % style
}
\makeatother

\makeatletter
\def\@fnsymbol#1{\ensuremath{\ifcase#1\or \textsuperscript{~\Letter}\or \ddagger\or
   \mathsection\or \mathparagraph\or \|\or **\or \dagger\dagger
   \or \ddagger\ddagger \else\@ctrerr\fi}}
\makeatother

\newcommand{\tableCellHeight}{1}
\newcommand{\tabstyle}[1]{
  \setlength{\tabcolsep}{#1}
  \renewcommand{\arraystretch}{\tableCellHeight}
  \centering
  \small
}

\definecolor{tabhighlight}{HTML}{e5e5e5}
\definecolor{citecolor}{HTML}{0071bc}
\usepackage[pagebackref=true,breaklinks=true,letterpaper=true,colorlinks,bookmarks=false]{hyperref}
\iccvfinalcopy % *** Uncomment this line for the final submission

\usepackage[capitalize]{cleveref}
\crefname{section}{Sec.}{Secs.}
\Crefname{section}{Section}{Sections}
\Crefname{table}{Table}{Tables}
\crefname{table}{Tab.}{Tabs.}

\def\iccvPaperID{****} % *** Enter the ICCV Paper ID here

\ificcvfinal\fi

\begin{document}

%%%%%%%%% TITLE
\title{Enhancing CLIP with GPT-4: Harnessing Visual Descriptions as Prompts}

% option1 : LLM's can prompt CLIP!?

\author{
    Mayug Maniparambil,
    Chris Vorster,
    Derek Molloy,
    Noel Murphy, \\
    Kevin McGuinness,
    Noel E. O'Connor \\
    ML Labs, Dublin City University,\\
    Dublin, Ireland
}

\maketitle
\ificcvfinal\thispagestyle{empty}\fi

\footnotetext[1]{Corresponding author: mayugmaniparambil@gmail.com}
%%%%%%%%% ABSTRACT
\begin{abstract}
   Contrastive pretrained large Vision-Language Models (VLMs) like CLIP have revolutionized visual representation learning by providing good performance on downstream datasets. VLMs are 0-shot adapted to a downstream dataset by designing prompts that are relevant to the dataset. Such prompt engineering makes use of domain expertise and a validation dataset. Meanwhile, recent developments in generative pretrained models like GPT-4 mean they can be used as advanced internet search tools. They can also be manipulated to provide visual information in any structure. In this work, we show that GPT-4 can be used to generate text that is visually descriptive and how this can be used to adapt CLIP to downstream tasks. We show considerable improvements in 0-shot transfer accuracy on specialized fine-grained datasets like EuroSAT (\(\sim 7\%\)), DTD (\(\sim 7\%\)), SUN397 (\(\sim 4.6\%\)), and CUB (\(\sim 3.3\%\))  when compared to CLIP's default prompt. We also design a simple few-shot adapter that learns to choose the best possible sentences to construct generalizable classifiers that outperform the recently proposed CoCoOP  by \(\sim 2\%\) on average and by over \(4\%\) on 4 specialized fine-grained datasets. The code, prompts, and auxiliary text dataset is available at \href{https://github.com/mayug/VDT-Adapter}{github.com/mayug/VDT-Adapter}.

\end{abstract}

%%%%%%%%% BODY TEXT
% TODO not sure where this fig should go:
\begin{figure*}[t]
\begin{center}
   \includegraphics[width=1.0\linewidth]{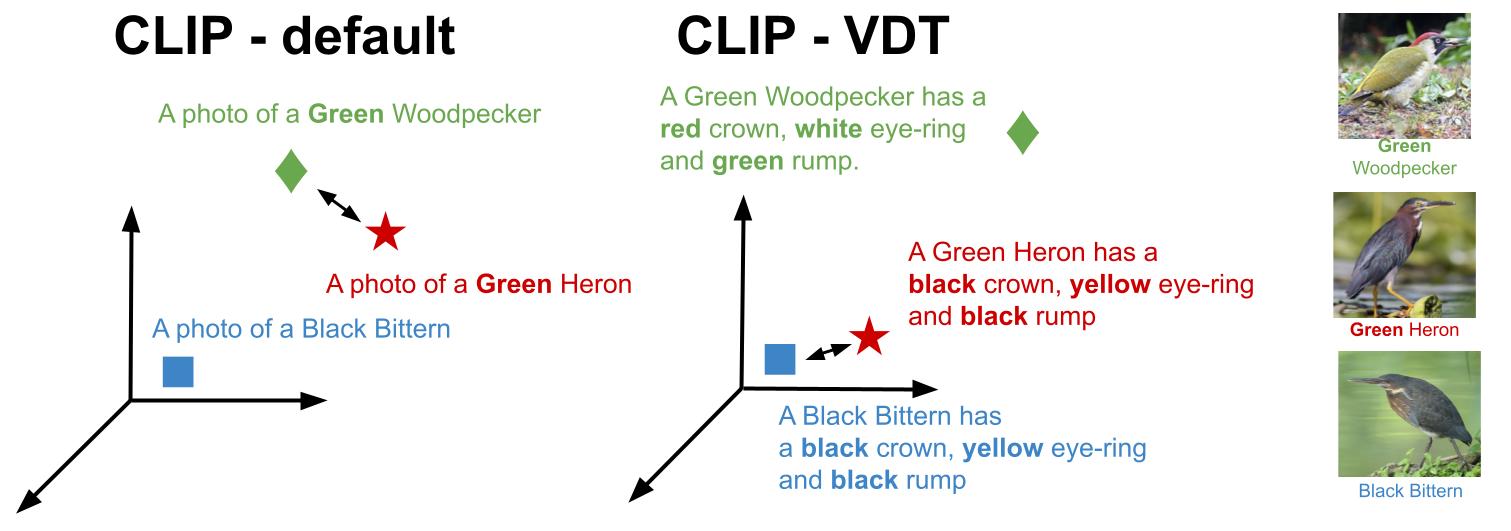}
\end{center}
   \caption{An example showing three birds, Green Heron, Green Woodpecker, and Black Bittern. Green Heron and Green Woodpecker have close-by classification prototypes by virtue of not having enough details in the prompt template. Only the text-encoder's embedding space is visualized. Here we see that adding visual descriptions to the prompt resolves this issue and moves the classification prototypes in the word-encoder's space such that classification prototypes for visually similar birds (Green Woodpecker and Black Bittern) lie together.}
\label{fig:illustrative}
\end{figure*}
\section{Introduction}

%%% Important point. Maybe write about how CLIP can be considered as a general visual  recognition system, and how for fine-grained recognition even humans require some descriptions of the thing we are trying to recognize. Hence we look at visual sentences...

Contrastive pre-training of large-scale VLMs has demonstrated remarkable image classification performance on open-set classes. Models like CLIP \cite{radford2021learning} and ALIGN \cite{Jia2021ScalingSupervision} are pretrained on web-scale datasets consisting of image-text pairs (over 400 million and 1.8 billion respectively), resulting in a highly generalizable model with competent 0-shot domain adaptation capabilities. While vanilla supervised training is performed on a closed set of concepts or classes, CLIP pretraining uses natural language. This results in a joint text-vision embedding space that is not constrained to a fixed set of classes. In CLIP, the classifier is constructed by plugging the class name into a predetermined prompt template like `a photo of \{class name\}'. A straightforward way to adapt CLIP to different domains is by prompt engineering, which usually involves modifying the prompt template to include semantic information about the target task. For example, to classify bird images, one could construct a prompt `a photo of \{classname\}, a type of bird'. This prompt engineering process, however, is not optimal because it:  1.) requires domain expertise in the target domain; 2.) has high variance -- small changes to the prompt result in large variation in performance; 3.) has a fixed prompt template for all the classes, therefore only the class name in the prompt provides the classification anchor, which might not contain enough information to distinguish different classes. For example, in Fig \ref{fig:illustrative} we see an image of a Green Heron, which from the name would suggest that it is predominantly a green-colored bird and we would assume that it is similar to Green Woodpecker if we have never seen either bird. However, we can see that it is in fact a blackish-brown bird with a chestnut-colored neck and visually more similar to a bird like the Black Bittern. For 0-shot transfer to fine-grained datasets like this to work well, CLIP has to either have seen and associated images of a Green Heron to the text `Green Heron' from its large pretraining dataset or additional information in the form of \textit{visually descriptive textual} (VDT) information is required. Here we define VDT as a set of sentences that describe the visual features of the class under consideration including shape, size, color, environment, patterns, composition, etc.  While most humans can identify many different common bird species just from their names, they would need access to an ornithology taxonomy of bird descriptions to identify more rare bird species. Similarly, we argue that CLIP's 0-shot accuracy can be improved by incorporating VDT information into the prompts. As shown, in Fig \ref{fig:illustrative}, including VDT information like \textit{black crown} and \textit{black rump} moves the classification prototype of Green Heron away from the  classification prototype of Green Woodpecker and towards that of Black Bittern in the text-encoder's embedding space.

In this work, we first show that we can use VDT information for each class in the target domain to construct class conditional prompts that achieve performance improvements over CLIP's default prompt. We show this on the CUB dataset \cite{Caltech-UCSD200} by constructing sentences from domain experts about the bird species in Section \ref{sec:vdt} as they are readily available as part of the dataset. 

However, we acknowledge that domain expert annotations are costly and time-consuming to obtain, hampering the scalability of our method to other datasets. To address this, we focus on the recent advances in \textit{generative pretrained Large Language Models (LLMs)} like GPT-4 to construct these class conditional prompts in a manner easily scalable to other datasets. These models are a good fit for the task of constructing sophisticated prompts, because: 1) they are a condensed form of human knowledge (trained on web-scale text data) \cite{YuKoLA:Models}; 2) they can be manipulated to produce information in any form or structure which makes compatibility with CLIP's prompt style relatively simple. Therefore we use GPT-4 to construct visually descriptive textual information about the classes with special emphasis in the GPT-4 prompts about visual cues like shape, color, structure, and compositionality. We use the generated VDT information to construct prompt ensembles that are passed through CLIP's text encoder and aggregated to generate classifiers that are then used for 0-shot classification. Using GPT-4 circumvents the need for domain knowledge and conveniently provides class conditional prompts. Prompt ensembling the VDT sentences reduce CLIP's performance sensitivity to small changes in the prompt.
We show performance improvements over vanilla CLIP with the default prompt on 12 datasets with an average improvement of 2\% and even better improvements in fine-grained datasets like EuroSAT  (\(\sim 7\%\)), DTD (\(\sim 7\%\)), SUN397 (\(\sim 4.6\%\)), and CUB (\(\sim 3.3\%\)). The prompts and all the auxiliary class information will be made publicly available to promote research in prompt ensembling and multi-modal adapter design.

Finally, we design a simple adapter that learns to adaptively select 
 and aggregate the best sentences for any given dataset and show that making use of this additional VDT information improves the few-shot domain transfer performance of CLIP as well. We demonstrate the few-shot adaptation performance for the recently proposed Base-to-New setting on a benchmark of 12 datasets and outperform recent methods like CoOp \cite{zhou2022learning} and CoCoOp \cite{zhou2022conditional} despite having fewer model parameters, shorter training time, and a simpler model architecture.

In short, our contributions are as follows:
\begin{enumerate}
    \item We show that including visually descriptive textual (VDT) information in prompts results in better 0-shot domain transfer performance of CLIP. 
    \item We use GPT-4 to generate VDT sentences in a scalable manner and show consistent performance improvements over CLIP in 0-shot domain transfer. 
    \item We design a simple adapter network to make use of this extra information for few-shot transfer and show performance improvements over methods like CLIP-Adapter and CoCoOp \cite{zhou2022conditional} for few-shot domain transfer in the Base-to-New setting.
    \item We release all the VDT information for all 12 datasets to promote further research in multi-modal prompt and adapter design for low-shot domain transfer of large VLMs. 
\end{enumerate}

%-------------------------------------------------------------------------

\section{Related Works}

\subsection{Vision Language Models}
% Historically vision and language models were generally constructed and trained independently, and their outputs often aligned using additional losses and modules. Image encoding often leveraged manually developed descriptors \cite{socher2013zero} or neural networks \cite{lei2015predicting, 7293699,frome2013devise}, while text encoding was accomplished via pre-trained word vectors \cite{socher2013zero, frome2013devise},  or frequency-grounded TF-IDF features \cite{elhoseiny2013write}. Various strategies such as metric learning \cite{frome2013devise}, multi-label classification \cite{joulin2016learning}, n-gram learning \cite{li2017learning}, and recently proposed image captioning \cite{desai2021virtex} were implemented for cross-modality alignment.

Recent VLMs \cite{Jia2021ScalingSupervision, radford2021learning, furst2022cloob} jointly learn the vision and language encoders from scratch and have demonstrated impressive 0-shot domain transfer performance. As mentioned in \cite{zhou2022learning}, this can be attributed to transformer networks \cite{vaswani2017attention}, contrastive losses \cite{chen2020simple, he2020momentum}, and web-scale training datasets \cite{radford2021learning, jia2021scaling}.

While our GPT-generated prompt ensembles are similar to CLIP's prompt ensembles, CLIP's prompt ensembles were constructed and tuned manually, and are class agnostic, while ours were generated by GPT models that were prompted to provide VDT information for each class.

\subsection{Prompt Learning}
% Prompt learning is a well-researched topic in NLP, where LLMs are viewed as a knowledge base and they are adapted to downstream tasks by only tuning a part of the prompt to the language model \cite{jiang2020can, shin2020autoprompt, zhong2021factual, li2021prefix, lester2021power, liu2021pre}.CoOp \cite{zhou2021coop} introduced prompt learning to VLM's by learning continuous prompts from a few-shot dataset, demonstrating impressive transfer performance to downstream datasets. However, \cite{zhou2022conditional} showed that CoOp often overfits to the few-shot dataset it is trained on, resulting in poor generalizability to new classes. They proposed CoCoOp, a prompt learning method that is image conditioned using a meta-network, demonstrating improvements over CoOp in the Base-to-New class setting. While this works, it is considerably more resource intensive than CoOp because of the image conditioning. Our work shows an orthogonal direction for the generalizability problem by making use of class conditional VDT information instead -- our CLIP-A-self, CLIP-A-mlp adapters improve upon CoCoOp despite having a simpler and less resource-hungry design.

CoOp \cite{zhou2022learning} successfully used prompt learning in VLMs but had generalizability limitations due to overfitting on the few-shot dataset \cite{zhou2022conditional}. In response, CoCoOp was proposed, enhancing performance with image-conditioned prompt learning using a meta-network, albeit at a higher resource cost. We address generalizability differently by using class conditional VDT information. Our simpler and more efficient model, CLIP-A-self, outperforms CoCoOp in the Base-to-New few-shot setting.

\subsection{Few-shot adapters for Vision Language models}
CLIP-Adapter \cite{gao2021clip} (CLIP-A) offers a simpler few-shot transfer method for VLMs, utilizing an MLP trained on fixed image/text encoders. Our CLIP-A-self is different from CLIP-A in that we apply a self-attention mechanism on the set of all sentences for any class, learning to select and aggregate the best subset of VDT information for the dataset from the few-shot training set. Although Tip-adapter \cite{zhang2022tip} showed superior performance on base classes with a cache model, it's inapplicable in the Base-to-New setting due to its reliance on few-shot test class examples, making it irrelevant for our comparison.
\subsection{Semantic information from Large Language Models}

Recent advancements in transformer-based language models, particularly the GPT family \cite{brown2020language, openai2023gpt4}, have demonstrated exceptional abilities in semantic extraction from intricate texts. Their application to vision tasks has emerged as an active area of research.
\cite{naeem2023i2mvformer} employs Palm540B LLM \cite{chowdhery2022palm} to generate semantic data for unsupervised class embedding vectors in 0-shot classification, but only tests on three legacy datasets. Our research presents results on a modern benchmark of 12 datasets. Recently, \cite{PrattWhatClassification, menon2022visual} leverage GPT-3 for class conditional prompts to enhance CLIP's 0-shot domain transfer on 6 datasets. While \cite{menon2022visual} focuses on using GPT-3 to construct visual descriptors that aid in the interpretability of CLIP's predictions during 0-shot domain transfer, we argue that 0-shot domain transfer performance improves with the inclusion of high-quality VDT information.  Hence, we make use of GPT-4 for richer, more diverse, and more accurate VDT information.

While \cite{menon2022visual} utilize GPT-3, probability space ensemble, and highlight VDT's role in 0-shot transfer, our method differs. We use GPT-4 for auxiliary data collection, perform ensemble in word-encoder space, and introduce a few-shot adapter for optimal VDT selection in few-shot transfer. \cite{UdandaraoSuS-X:Models} uses GPT-3 for prompt construction in diffusion models to generate images for support sets while our work only uses GPT4 to acquire auxiliary text data. To our knowledge, we are the first to prompt GPT-4 for visually descriptive sentences to improve CLIP's 0-shot and few-shot domain transfer.

\section{Methodology}
\subsection{Review of CLIP and CLIP-Adapter}

Through contrastive pretraining on large image-text datasets, CLIP performs image classification on various concepts, aligning related images and texts in a shared embedding space, while separating dissimilar ones.  After pretraining, CLIP directly performs image classification on the target dataset without any finetuning. First, we review how the CLIP model performs 0-shot classification on an open set.

The CLIP model, comprising a vision and language model, encodes an image and its corresponding caption into visual and textual embeddings, respectively. During inference, these embeddings are compared using \textit{cosine similarity}. Given an image \(I\in \mathbb{R}^{H\times W \times C}\), where $H$, $W$, $C$ denotes the height, width, and number of channels of the image, the vision encoder transforms the image into the joint embedding space to get the image features \(f\in \mathbb{R}^{D}\) where $D$ represents the dimension of the features.

During inference, a prompt template such as `A photo of \{classname\}' is used to generate sentences for $K$ different classes and passed through the text-encoder to yield classifier weight matrix \(W\in \mathbb{R}^{D \times K}\). Prediction probabilities are then calculated by multiplying image feature $f$ with $W$ and applying a softmax function: 

\begin{equation}
    \label{logits}
    % \begin{align}
        f = \mathrm{Backbone}(\mathbf{I}),~~p_i = \frac
        {\operatorname{exp}(\mathbf{W}_{i}^T f) / \tau}
        {\sum_{j=1}^{K} \operatorname{exp}(\mathbf{W}_{j}^T f) / \tau}, 
    % \end{align}
\end{equation}

% In CLIP \cite{radford2021learning}, 0-shot domain transfer is achieved by appending domain-specific information into the prompt template. For instance, for photos of birds, a prompt template like `A photo of a \{class-name\}, a type of bird' is used to generate the classification vectors. The work in \cite{radford2021learning} reports that careful prompt design and prompt ensembling are important to improve 0-shot classification accuracy. Prompt ensembling is achieved by constructing several prompts for each class and then averaging the classification vectors. In our work, we show that prompt ensembles of VDT information improve CLIP's 0-shot domain transfer.

In CLIP \cite{radford2021learning}, 0-shot domain transfer utilizes domain-specific information in the prompt template, such as `A photo of a \{class-name\}, a type of bird' for bird images. \cite{radford2021learning} reports that careful prompt design and prompt ensembling are important to improve 0-shot classification accuracy. Prompt ensembling is achieved by constructing several prompts for each class and then averaging the classification vectors. In our work, we show that prompt ensembles of VDT information improve CLIP's 0-shot domain transfer.
%%Unlike \cite{RadfordLearningSupervision} which makes use of domain knowledge and trial and error to design the prompts, in this paper we make use of GPT models to construct the prompt ensembles.    %% something about our method here..

% CLIP-A \cite{gao2021clip} is a simple learnable MLP adapter on top of the image encoder features and/or word encoder features to enable few-shot transfer to the target datasets. During few-shot transfer, we are given $N$ images for each class of the base dataset along with their labels. Let \(\left ( x_{i,k},  y_{i,k}\right)_{i=1,k=1}^{i=N,j=K}\) be the image, label pairs comprising the few-shot training set. CLIP-A constructs $K$ classifier weights using the prompt template $H$ to get the classifier weight \(W\in \mathbb{R}^{D \times K}\) by plugging in the class name \(classname(\left \{ y_{i,k} \right \}) \) into the prompt template $H$ and transforming using the text encoder $g$ as \(W = g(H(classname(\left \{ y_{i,k} \right \}) ))\). Then image features $f$ and text features $W$ are passed through the learnable adapters \(A_{v}\), \(A_{t}\) respectively to obtain the adapted features as follows:

CLIP-A \cite{gao2021clip} is a learnable MLP adapter applied to image and/or word encoder features for few-shot transfer to target datasets. During few-shot transfer, given $N$ images per class with labels, denoted as \(\left ( x_{i,k},  y_{i,k}\right)_{i=1,k=1}^{i=N,j=K}\), $K$ classifier weights are constructed using the prompt template $H$ and text encoder $g$ as \(W = g(H(classname(\left \{ y_{i,k} \right \}) ))\). The image features $f$ and text features $W$ pass through the learnable adapters \(A_{v}\), \(A_{t}\) to get adapted features as follows.

\begin{align}
        f^{\star} &= \alpha A_{v}(f)^{T} + (1-\alpha) f, \\
        \mathbf{W}^{\star} &= \beta A_{t}(\mathbf{W})^{T} + (1 - \beta) \mathbf{W}.
\end{align}

The hyperparameters \(\alpha\) and \(\beta\) blend CLIP's knowledge with fine-tuned knowledge to avoid CLIP-Adapter overfitting. Logits are calculated as per Eqn \ref{logits}, and cross entropy loss over the entire training set \(\left ( x_{i,k},  y_{i,k}\right)_{i=1,k=1}^{i=N,j=K}\) is used to optimize \(A_{v}\), \(A_{t}\).

In the \textit{All} setting, few-shot transfer is tested on a hold-out dataset with images from the $K$ classes used in training. In the Base-to-New setting, proposed by \cite{zhou2022conditional}, the evaluation occurs on $U$ non-overlapping classes. Our model is evaluated in the more practical Base-to-New setting.

% From clip adapter paper we make use of the variant of the CLIP-Adapter with only visual adaptation for all our comparisons. # write about this in baselines section.

\begin{figure*}[t]
\begin{center}
   \includegraphics[width=1.0\linewidth]{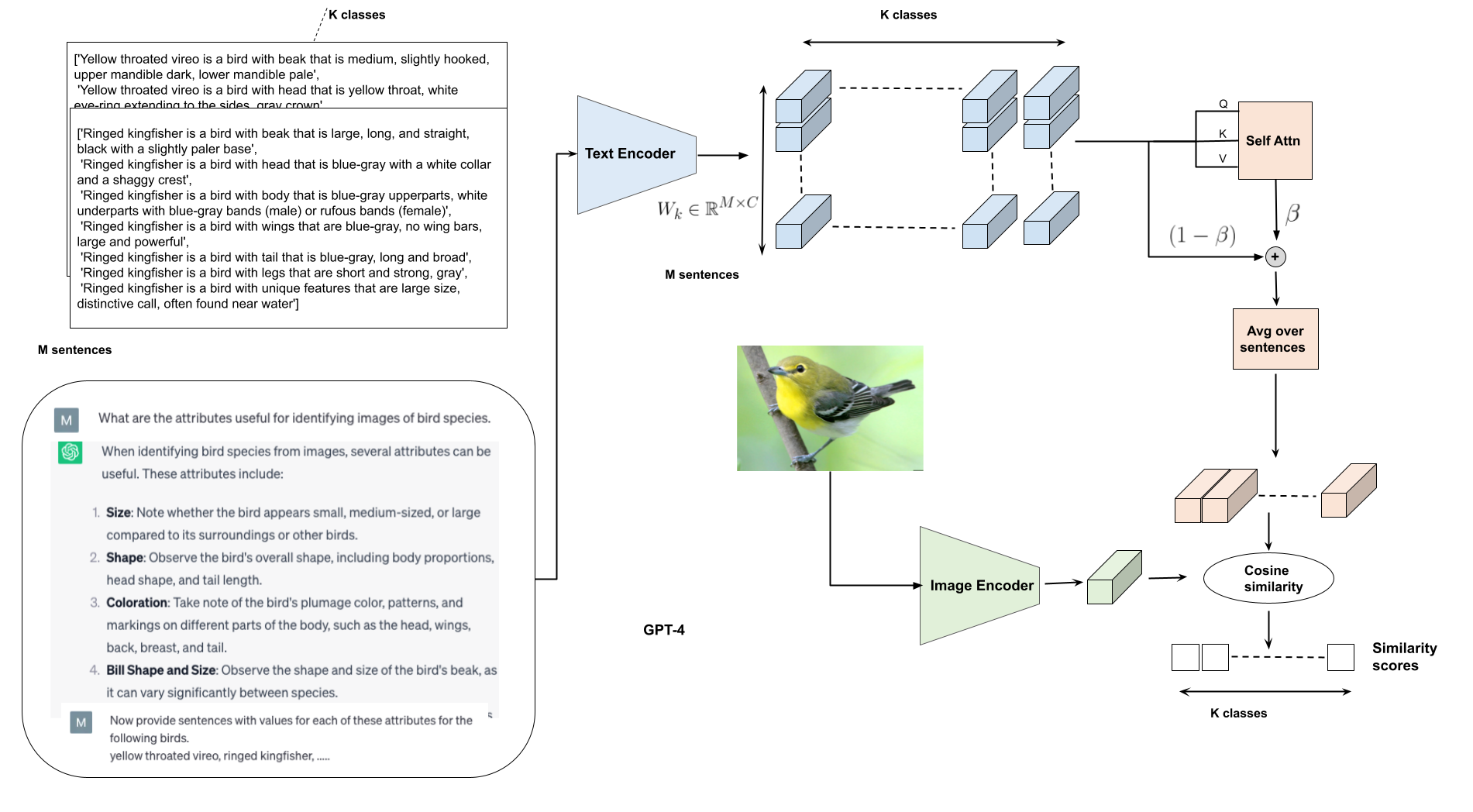}
\end{center}
   \caption{CLIP-A-self, our simple self-attention based adapter learns to select and aggregate the most relevant subset of Visually Descriptive Text (VDT) to generate more generalizable classifiers. First, we prompt GPT-4 to generate VDT, N sentences for K classes that are then passed through the text encoder to get embeddings for each of the N*K sentences. Self-attention is applied over the N sentences of each class and averaged to get K adapted classifier embeddings. }
\label{fig:few-shot}
\end{figure*}

\subsection{Language Model Prompt Design}

In this section, we show that using VDT information in the prompt template improves CLIP's 0-shot transfer capabilities and describe our approach to generate class-specific prompts using an LLM.

\subsubsection{Visual Descriptive Sentences} \label{sec:vdt}
\cite{radford2021learning} demonstrates that careful prompt design and prompt ensembling improve the 0-shot classification performance of CLIP. Here we ask the question: What type of information can be appended to the prompt template to improve the 0-shot domain transfer performance? We show that appending visually descriptive information to the prompt template and ensembling improves the 0-shot performance over the default prompt and prompts containing non-visual information.

Using the CUB dataset with expert annotations, we contrast the 0-shot performance of visual and non-visual prompt ensembles. For the visual prompts, we take class attribute vectors detailing attributes like color, pattern, shape, etc. for 28 bird body parts, leading to 312 scores per bird. We use the most pronounced attribute-value pairs to form 28 visual prompts (denoted \textit{Visual-GT}) such as `A photo of Green Heron. Green Heron has a greenish-black head cap.' Conversely, for non-visual prompts (denoted \textit{Non-Visual-GT}), we collect information on bird calls, migration, behavior, and habitat, yielding 12  different prompts like `A photo of Green Heron. The green heron's bird call is a loud, harsh `skeow'' per class.

% We ensemble by averaging the class-level sentence embeddings in CLIP's joint embedding space to get the classification vectors for both the visual set and the non-visual set. This is because CLIP's context space is only 77 tokens long and all the visual information cannot be appended to the prompt template to create one large sentence. In table \ref{table_visual}, we show that the prompt with non-visual information provides no further improvement over the default prompt, while the prompt with visual information provides 4\(\%\) improvement over CLIP. %% TODO: para-shorten

We derive classification vectors for \textit{Visual-GT} and \textit{Non-Visual-GT} by averaging class-level sentence embeddings within CLIP's joint embedding space, considering its 77-token limit. Table \ref{table_visual} shows no improvement using \textit{Non-Visual-GT} prompts over the default, yet a 4\(\%\) improvement with \textit{Visual-GT}.

%%% any other way to show that the classifiers are superior?
%%% t-sne plots?

% \begin{table}[]
% \caption{Comparing visual and non-visual prompt ensembles for 0-shot domain transfer to the CUB dataset.}
% \begin{tabular}{llll}
% \toprule
% Method                                                                      & Base & New  & H \\
% \midrule
% \vspace{1em}
% \begin{tabular}[c]{@{}l@{}}CLIP \\ default prompt\end{tabular}              & 58.7 & 70.3 & 63.97         \\
% \vspace{1em}
% \begin{tabular}[c]{@{}l@{}}CLIP \\ default prompt + non-visual-gt\end{tabular} & 57.5 & 71.0 & 63.54         \\
% \vspace{1em}
% \begin{tabular}[c]{@{}l@{}}CLIP \\ default prompt + visual-gt\end{tabular}     & \textbf{62.3} & \textbf{73.9} & \textbf{67.60}   \\
% \vspace{1em}
% \begin{tabular}[c]{@{}l@{}}CLIP \\ default prompt + visual-gpt\end{tabular}     & \textbf{62.0} & \textbf{73.1} & \textbf{67.09} \\
% \bottomrule
% \end{tabular}
% \label{table_visual}
% \end{table}

\begin{table}[]
\caption{Comparing visual and non-visual prompt ensembles for 0-shot domain transfer to the CUB dataset.}
% \begin{tabular}{llll}
% \toprule
% Method                                                                      & Base & New  & H \\
% \midrule
% % \vspace{1em}
% \begin{tabular}[c]{@{}l@{}}default prompt\end{tabular}              & 58.7 & 70.3 & 63.97         \\
% % \vspace{1em}
% \begin{tabular}[c]{@{}l@{}}non-visual-gt\end{tabular} & 57.5 & 71.0 & 63.54         \\
% % \vspace{1em}
% \begin{tabular}[c]{@{}l@{}}visual-gt\end{tabular}     & \textbf{62.3} & \textbf{73.9} & \textbf{67.60}   \\
% % \vspace{1em}
% \begin{tabular}[c]{@{}l@{}} visual-gpt\end{tabular}     & \textbf{62.0} & \textbf{73.1} & \textbf{67.09} \\
% \bottomrule
% \end{tabular}
\centering
\resizebox{\columnwidth}{!}{
\begin{tabular}{lllll}
\toprule
Prompting & Default & \makecell[c]{Non-Visual-\\GT} & \makecell[c]{Visual-\\GT} & \makecell[c]{Visual-\\GPT} \\
\midrule
Accuracy  & \makecell[c]{54.7}    &  \makecell[c]{53.0}  & \makecell[c]{57.7} & \makecell[c]{57.4} \\
\bottomrule
\end{tabular}
}%
\label{table_visual}
\end{table}

\begin{table}[t]
    \caption{Results of including LLM generated VDT on 6 datasets for comparison with other works. We see that higher quality VDT from GPT-4 outperforms GPT-3 generated VDT on specialized datasets like DTD OxfordPets and EuroSAT.}
    \centering
    \tabcolsep=0.8mm
    % \footnotesize
    \resizebox{\columnwidth}{!}{%
    \begin{NiceTabular}{ccccccc|c}
        \toprule
        Methods & EuroSAT & Food101  & DTD & \makecell[c]{Oxford\\Pets} & CUB & ImageNet & Average \\
        \midrule
        CLIP & 47.69 & 85.97 & 43.09 & 89.07 & 54.70 &64.51 & 64.17 \\
        DCLIP\cite{menon2022visual} & 48.82 & \textbf{88.50} & 45.59 & 86.92 & \textbf{57.75} & 68.03 & 65.93 \\
        CLIP-GPT & \textbf{54.86} & 86.43 & \textbf{50.15} & \textbf{91.54} & 57.43& \textbf{68.92} &  \textbf{68.21} \\

        \bottomrule
    \end{NiceTabular}
    }%
    \label{table_subset_results_zs}
\vspace{-5mm}
\end{table}
\begin{table*}[t]
    \caption{Results of 12 datasets with ViT-B/16.}
    \centering
    \tabcolsep=0.8mm
    % \footnotesize
    \resizebox{\textwidth}{!}{%
    \begin{NiceTabular}{ccccccccccccc|c}
        \toprule
        Methods & EuroSAT & Caltech101 & \makecell[c]{Oxford\\Flowers} & Food101 & \makecell[c]{FGVC\\Aircraft} & DTD & \makecell[c]{Oxford\\Pets} & \makecell[c]{Stanford\\Cars} & Sun397 & UCF101 & CUB & ImageNet & Average \\
        \midrule
        CLIP & 47.69 & 93.75 & 70.69 & 85.97 & \textbf{24.81} & 43.09 & 89.07 & \textbf{65.55} & 62.61 & \textbf{67.54} & 54.70 &64.51 & 64.16 \\
        CLIP-GPT & \textbf{54.86} & \textbf{94.51} & \textbf{73.40} & \textbf{86.43} & 23.42 & \textbf{50.15} & \textbf{91.54} & 65.01 & \textbf{67.24} & 65.51 & \textbf{57.43} & \textbf{68.9} & \textbf{66.53} \\

        \bottomrule
    \end{NiceTabular}
    }%
    \label{table_main_results_zs}
\vspace{-2mm}
\end{table*}

\subsubsection{Prompting LLMs for visually descriptive information}

% In the previous section, we show that visually descriptive information from domain experts can be used to construct class conditional prompts that improve 0-shot performance. However, obtaining domain expert annotations is expensive and time-consuming. GPT language models are a good source of knowledge and their knowledge generally scales with size \cite{YuKoLA:Models}. They can also be easily manipulated to generate output in any manner that we want by simply modifying the prompt. Hence we make use of a simple strategy to prompt GPT-4 to obtain visually descriptive information for any given dataset that can then be used to construct prompt ensembles for CLIP.

In the prior section, we highlighted the use of expert VDT information in creating class-specific prompts to enhance CLIP's 0-shot performance. However, acquiring expert annotations is both expensive and time-consuming. To overcome this, we utilize GPT language models, known for their large-scale knowledge and flexibility \cite{YuKoLA:Models}. Our approach involves using GPT-4 to generate visual descriptions for any given dataset thereby aiding in the construction of prompt ensembles for CLIP in a scalable manner.

Our prompting strategy takes inspiration from chain-of-thought prompting \cite{wei2022chain} and is as follows: First, we ask GPT-4 to list all the attributes that may be necessary to discriminate between images of the $K$ classes under consideration. Second, we ask GPT-4 to provide the values for all these attributes for all the $K$ classes as sentences. An example for the CUB dataset is shown in the left side of Fig \ref{fig:illustrative}.

The last row in Table \ref{table_visual} shows that the GPT-4 generated visual sentences' performance is similar to that of sentences generated from the class attribute vectors annotated by domain experts. We follow the same simple strategy for all the datasets in the benchmark suite to generate visually descriptive sentences in a scalable and flexible manner and use them to construct prompt ensembles.

\subsection{Simple few-shot adapters for visual sentences}

\begin{table}
    \centering
    \begin{tabular}{l cc|c}
    \toprule
    & Base & New & H \\
    \midrule
    CLIP & 68.45 & 73.89 & 71.05 \\
    CoOp & \textbf{82.39} & 62.39 & 70.99 \\
    CoCoOp & 79.35 & 71.89 & 75.37 \\
    CLIP-A &  78.90 & 72.14 & 75.07 \\
    \rowcolor{tabhighlight}
    % CLIP-A-mlp &  80.06 & \textbf{75.24} & 77.42 \\
    \rowcolor{tabhighlight}
    CLIP-A-self &  82.12 & 74.20 & \textbf{77.78} \\
    \bottomrule
    \end{tabular}
    \label{table_adapter_avg}
    \caption{Comparing our CLIP-A-self against other methods on average accuracy over 12 datasets.}
\end{table}

%% Redoing and removing mlp adapter completely.

We design a simple adapter that can use VDT information to improve the few-shot transfer of CLIP to the target datasets. Similar to the CLIP-A text, we append a small set of learnable parameters to the output of the word encoder and train the adapter using cross-entropy loss. Our CLIP-A-self uses a self-attention layer that applies attention over the embeddings of the different sentences for each class and averages the output to get the final classification vector.

Given we have $M$ GPT generated sentences for each of the $K$ classes \(t_{m,k}\), we construct $M$ prompts by appending each sentence to the prompt template like  \(H(classname(y_{i,k} ), \left \{ t_{m,k} \right \})\) and pass them through CLIP's word encoder to get \(W^{sent}\in \mathbb{R}^{D \times M \times K}\). 

% Then for CLIP-A-mlp we use a 3-layer MLP \(B_{mlp}\) as follows:

% \begin{align}
%     W_{avg} = 1/M \sum _{m=1}^{M} W_{m,k}^{sent}  \\
%     W^{\star} = \beta B_{mlp}(\mathbf{W_{avg}})^{T} + (1 - \beta) \mathbf{W_{avg}}
% \end{align}

For the self-attention adapter, we apply vanilla self-attention \cite{vaswani2017attention} over all the visual descriptive sentences such that during training it learns to select and aggregate the most relevant visual sentences for identifying each class. Just like before, we first obtain the classification vector for all sentences \(W^{s}\in \mathbb{R}^{K \times M \times D}\) and pass them as the key, query, and value to the self-attention module \(B_{self}\) and average out the output tokens to get the final classification vector \(W^{\star}\). Here the attention is applied over the $M$ different visually descriptive sentences. 
\begin{align}
    &W_{avg} = 1/M \sum _{m=1}^{M} W_{m,k}^{s}  \\
    &\left \{W^{a}_{m,k}\right \}_{1}^{M} = B_{self}(\left \{ W^{s}_{m,k} \right \}_{1}^{M}, \left \{ W^{s}_{m,k} \right \}_{1}^{M}, \left \{ W^{s}_{m,k} \right \}_{1}^{M})\\
    &W_{a-mean} = 1/M \sum _{m=1}^{M} W_{m,k}^{a}  \\
    &W^{\star} = \beta \mathbf{W_{a-mean}}^{T} + (1 - \beta) \mathbf{W_{avg}}
\end{align}

 We finally obtain the new adapter classifier weights \(W^{\star}\in \mathbb{R}^{D \times K}\) that have been adapted to focus on the most visually discriminative information among the $M$ visually descriptive sentences for any given dataset. We make use of \ref{logits} to calculate the probabilities and predict the image category by selecting the class with the highest probability.

During the few-shot training only the weights of the adapter network  \(B_{self}\) are trained using cross-entropy loss.

\begin{table*}[t]
    \tabstyle{6pt}
    \caption{\textbf{Comparison of GPT-Adapters with CLIP, CoOp and CoCoOp in the Base-to-New generalization setting}. For prompt learning-based methods (CoOp and CoCoOp), their prompts are learned from the base classes (16 shots). The results strongly justify the importance of including extra visual information. H denotes Harmonic mean (to highlight the generalization trade-off~\cite{xian2018zero}).}
    \label{table_results_generalization}
    % \begin{subtable}[t]{.3\textwidth}
    % \centering
    % \caption{\textbf{Average over 12 datasets}.}
    % \begin{tabular}{l cc|c}
    % \toprule
    % & Base & New & H \\
    % \midrule
    % CLIP & 68.45 & 73.89 & 71.05 \\
    % CoOp & \textbf{82.39} & 62.39 & 70.99 \\
    % CoCoOp & 79.35 & 71.89 & 75.37 \\
    % CLIP-A &  78.90 & 72.14 & 75.07 \\
    % \rowcolor{tabhighlight}
    % % CLIP-A-mlp &  80.06 & \textbf{75.24} & 77.42 \\
    % \rowcolor{tabhighlight}
    % CLIP-A-self &  82.12 & 74.20 & \textbf{77.78} \\
    % \bottomrule
    % \end{tabular}
    % \end{subtable}
    \vspace{1em}  
    \begin{subtable}[t]{.3\textwidth}
    \centering
    \caption{CUB.}
    \begin{tabular}{l cc|c}
    \toprule
    & Base & New & H \\
    \midrule
    CLIP &  58.7 & 70.3 & 63.90 \\
    CoOp & 79.2 & 53.3 & 63.71 \\
    CoCoOp & 67.1 & 74.1 & 70.40 \\
    CLIP-A &  68.3 & 70.8 & 69.53 \\
    \rowcolor{tabhighlight}
    % CLIP-A-mlp &  74.1 & 72.8 & 73.44 \\
    \rowcolor{tabhighlight}
    CLIP-A-self &  \textbf{78.6} & \textbf{71.3} & \textbf{74.77} \\
    \bottomrule
    \end{tabular}
    \end{subtable}
    ~
    \begin{subtable}[t]{.3\textwidth}
    \centering
    \caption{Caltech101.}
    \begin{tabular}{l cc|c}
    \toprule
    & Base & New & H \\
    \midrule
    CLIP & 96.84 & 94.00 & 95.40 \\
    CoOp & 98.00 & 89.81 & 93.73 \\
    CoCoOp & 97.96 & 93.81 & 95.84 \\
    CLIP-A &  97.7 & 93.6 & 95.61 \\
    \rowcolor{tabhighlight}
    % CLIP-A-mlp &  97.5 & 95.7 & 96.59 \\
    \rowcolor{tabhighlight}
    CLIP-A-self &  \textbf{98.3} & \textbf{95.9} & \textbf{97.09} \\
    \bottomrule
    \end{tabular}
    \end{subtable}
    ~
    \begin{subtable}[t]{.3\textwidth}
    \centering
    \caption{OxfordPets.}
    \begin{tabular}{l cc|c}
    \toprule
    & Base & New & H \\
    \midrule
    CLIP & 91.17 & 97.26 & 94.12 \\
    CoOp & 93.67 & 95.29 & 94.47 \\
    CoCoOp & \textbf{95.20} & \textbf{97.69} & \textbf{96.43} \\
    CLIP-A &  94.8 & 97.0 & 95.89 \\
    \rowcolor{tabhighlight}
    % CLIP-A-mlp &  95.5 & 96.9 & 96.19 \\
    \rowcolor{tabhighlight}
    CLIP-A-self &  94.4 & 97.0 & 95.68 \\
    \bottomrule
    \end{tabular}
    \end{subtable}
    \vspace{1em}
    \begin{subtable}[t]{.3\textwidth}
    \centering
    \caption{StanfordCars.}
    \begin{tabular}{l cc|c}
    \toprule
    & Base & New & H \\
    \midrule
    CLIP & 63.37 & \textbf{74.89} & 68.65 \\
    CoOp & \textbf{78.12} & 60.40 & 68.13 \\ 
    CoCoOp & 70.49 & 73.59 & 72.01 \\
    CLIP-A &  70.5 & 73.3 & 71.87 \\
    \rowcolor{tabhighlight}
    % CLIP-A-mlp &  72.2 & 73.1 & 72.65 \\
    \rowcolor{tabhighlight}
    CLIP-A-self &  76.8 & 72.9 & \textbf{74.80} \\
    \bottomrule
    \end{tabular}
    \end{subtable}
    ~
    \begin{subtable}[t]{.3\textwidth}
    \centering
    \caption{Flowers102.}
    \begin{tabular}{l cc|c}
    \toprule
    & Base & New & H \\
    \midrule
    CLIP & 72.08 & \textbf{77.80} & 74.83 \\
    CoOp & \textbf{97.60} & 59.67 & 74.06 \\
    \rowcolor{tabhighlight}
    CoCoOp & 94.87 & 71.75 & 81.71 \\
    CLIP-A &  94.6 & 71.5 & 81.44 \\
    \rowcolor{tabhighlight}
    % CLIP-A-mlp &  98.1 & 76.2 & \textbf{85.77} \\
    \rowcolor{tabhighlight}
    CLIP-A-self &  97.4 & 75.3 & \textbf{84.94} \\
    \bottomrule
    \end{tabular}
    \end{subtable}
    ~
    \begin{subtable}[t]{.3\textwidth}
    \centering
    \caption{Food101.}
    \begin{tabular}{l cc|c}
    \toprule
    & Base & New & H \\
    \midrule
    CLIP & 90.10 & 91.22 & 90.66 \\
    CoOp & 88.33 & 82.26 & 85.19 \\
    CoCoOp & \textbf{90.70} & \textbf{91.29} & \textbf{90.99} \\
    CLIP-A &  90.3 & 91.2 & 90.75 \\
    \rowcolor{tabhighlight}
    % CLIP-A-mlp &  90.6 & 91.6 & 91.10 \\
    \rowcolor{tabhighlight}
    CLIP-A-self &  90.4 & 91.2 & 90.80 \\
    \bottomrule
    \end{tabular}
    \end{subtable}
    \vspace{1em}
    \begin{subtable}[t]{.3\textwidth}
    \centering
    \caption{FGVCAircraft.}
    \begin{tabular}{l cc|c}
    \toprule
    & Base & New & H \\
    \midrule
    CLIP & 27.19 & \textbf{36.29} & 31.09 \\
    CoOp & \textbf{40.44} & 22.30 & 28.75 \\
    CoCoOp & 33.41 & 23.71 & 27.74 \\
    CLIP-A &  34.9 & 33.5 & 34.19 \\
    \rowcolor{tabhighlight}
    % CLIP-A-mlp &  35.0 & 35.0 & 35.00 \\
    \rowcolor{tabhighlight}
    CLIP-A-self &  37.8 & 33.0 & \textbf{35.24} \\
    \bottomrule
    \end{tabular}
    \end{subtable}
    ~
    \begin{subtable}[t]{.3\textwidth}
    \centering
    \caption{SUN397.}
    \begin{tabular}{l cc|c}
    \toprule
    & Base & New & H \\
    \midrule
    CLIP & 69.36 & 75.35 & 72.23 \\
    CoOp & \textbf{80.60} & 65.89 & 72.51 \\
    % \rowcolor{tabhighlight}
    CoCoOp & 79.74 & 76.86 & 78.27 \\
    CLIP-A &  80.1 & 75.9 & 77.94 \\
    % \rowcolor{tabhighlight}
    % CLIP-A-mlp &  80.2 & \textbf{78.4} & \textbf{79.29} \\
    \rowcolor{tabhighlight}
    CLIP-A-self &  81.4 & 76.8 & 79.03 \\
    \bottomrule
    \end{tabular}
    \end{subtable}
    ~
    \begin{subtable}[t]{.3\textwidth}
    \centering
    \caption{DTD.}
    \begin{tabular}{l cc|c}
    \toprule
    & Base & New & H \\
    \midrule
    CLIP & 53.24 & 59.90 & 56.37 \\
    CoOp & 79.44 & 41.18 & 54.24 \\
    CoCoOp & 77.01 & 56.00 & 64.85 \\
    CLIP-A &  74.9 & 53.0 & 62.08 \\
    \rowcolor{tabhighlight}
    % CLIP-A-mlp &  75.8 & 62.0 & 68.21 \\
    \rowcolor{tabhighlight}
    CLIP-A-self &  \textbf{81.8} &\textbf{62.3} & \textbf{70.73} \\
    \bottomrule
    \end{tabular}
    \end{subtable}
    ~
    \begin{subtable}[t]{.3\textwidth}
    \centering
    \caption{EuroSAT.}
    \begin{tabular}{l cc|c}
    \toprule
    & Base & New & H \\
    \midrule
    CLIP & 56.48 & 64.05 & 60.03 \\
    CoOp & \textbf{92.19} & 54.74 & 68.69 \\
    CoCoOp & 87.49 & 60.04 & 71.21 \\
    CLIP-A &  82.5 & 62.4 & 71.06 \\
    \rowcolor{tabhighlight}
    % CLIP-A-mlp &  85.7 & \textbf{72.6} & \textbf{78.61} \\
    \rowcolor{tabhighlight}
    CLIP-A-self &  88.5 & 70.5 & 78.48 \\
    \bottomrule
    \end{tabular}
    \end{subtable}
    ~
    \begin{subtable}[t]{.3\textwidth}
    \centering
    \caption{UCF101.}
    \begin{tabular}{l cc|c}
    \toprule
    & Base & New & H \\
    \midrule
    CLIP & 70.53 & \textbf{77.50} & 73.85 \\
    CoOp & \textbf{84.69} & 56.05 & 67.46 \\
    CoCoOp & 82.33 & 73.45 & 77.64 \\
    CLIP-A &  82.9 & 74.9 & 78.70 \\
    \rowcolor{tabhighlight}
    % CLIP-A-mlp &  80.2 & 78.6 & 79.39 \\
    \rowcolor{tabhighlight}
    CLIP-A-self &  84.1 & 76.4 & \textbf{80.07} \\
    \bottomrule
    \end{tabular}
    \end{subtable}
    ~
    \begin{subtable}[t]{.3\textwidth}
    \centering
    \caption{ImageNet.}
    \begin{tabular}{l cc|c}
    \toprule
    & Base & New & H \\
    \midrule
    CLIP & 72.43 & 68.14 & 70.22 \\
    CoOp & \textbf{76.47} & 67.88 & 71.92\\
    % \rowcolor{tabhighlight}
    CoCoOp & 75.98 & \textbf{70.43} & \textbf{73.10} \\
    CLIP-A &  75.4 & 68.6 & 71.84 \\
    \rowcolor{tabhighlight}
    % CLIP-A-mlp &  75.9 & 75.0 & 72.83 \\
    \rowcolor{tabhighlight}
    CLIP-A-self &  76.4 & 68.3 & 72.12 \\
    \bottomrule
    \end{tabular}
    \end{subtable}

\end{table*}

\begin{table}[]
\centering
% \resizebox{\columnwidth}{!}{%
\begin{tabular}{l|llll}
\toprule
Prompting      & ZS   & \begin{tabular}[c]{@{}l@{}}Base\end{tabular} & \begin{tabular}[c]{@{}l@{}}New\end{tabular} & \begin{tabular}[c]{@{}l@{}}H\end{tabular} \\
\midrule
Default        & 54.7 & NA                                                        & NA                                                       & NA                                                     \\
OpenAssistant & 56.0 & 78.3                                                      & 69.8                                                     & 73.80                                                  \\
GPT-3.5        & 55.7 & 78.1                                                      & 70.6                                                     & 74.16                                                  \\
GPT-4          & 57.4 & 78.6                                                      & 71.3                                                     & 74.77                                                 
\end{tabular}%
% }
\caption{Comparing different GPT models for obtaining the VDT information. We see that the larger models provide higher quality VDT information but CLIP-A-self is capable of producing generalizable classifiers even with smaller models like OpenAssistant.}
\label{table:gpt_models}
\end{table}

\section{Experiments}
% We evaluate the importance of visual sentence ensembles in the following two problem settings: (i) We evaluate the quality of the visual sentences themselves by comparing an ensemble of visual sentence prompts with that of the default CLIP prompt for each dataset on a benchmark suite of 12 datasets; (ii)  We compare the performance of adapters that make use of these visual sentence prompts with other few-shot transfer methods on generalization from Base-to-New classes within a dataset. Before discussing the results we provide information about the datasets and experimental setup.

We assess the significance of visual sentence ensembles in two scenarios: (i) we gauge visual sentence quality by comparing an ensemble of these prompts with CLIP's default prompts across 12 benchmark datasets; (ii) we contrast the performance of adapters using these visual prompts against other few-shot transfer techniques in Base-to-New class generalization within a dataset. Prior to discussing the results, we detail the datasets and experimental setup.

\subsection{Datasets}

We use 11 diverse image recognition datasets from \cite{zhou2022learning} and the bird species CUB dataset \cite{Caltech-UCSD200} for both study settings, extending our suite to 12. These include generic object datasets ImageNet~\cite{deng2009imagenet} and Caltech101~\cite{fei2004learning}; fine-grained classification datasets OxfordPets~\cite{parkhi2012cats}, StanfordCars~\cite{krause20133d}, Flowers102~\cite{nilsback2008automated}, Food101~\cite{bossard2014food} and FGVCAircraft~\cite{maji2013fine}; SUN397~\cite{xiao2010sun} for scene recognition; UCF101~\cite{soomro2012ucf101} for action recognition; DTD~\cite{cimpoi2014describing} for texture classification; EuroSAT~\cite{helber2019eurosat} for satellite imagery; and CUB for bird identification. 

For 0-shot transfer with visual sentences, we test on  \emph{All} classes across these datasets while for the Base-to-New setting, following \cite{zhou2022conditional}, we equally sample classes for base and new sets without overlap. We use the 150-base and 50-new class split from ZSL and few-shot literature \cite{xian2018zero, maniparambil2022} for CUB. Like \cite{zhou2022conditional}, our CLIP-A-self is evaluated on the 16-shot setting for easier comparison with other methods.

\subsection{Baselines}
We compare the performance of visual sentences ensemble on 0-shot transfer against the CLIP model \cite{radford2021learning} whose default prompts for each dataset have been extensively fine-tuned using a test set. We also compare against DCLIP \cite{menon2022visual} a recent work that uses GPT-3 to generate VDT information for 0-shot transfer.
We compare our CLIP-A-self against two prompt learning methods CoOp \cite{zhou2022learning} which learns static prompts and CoCoOp \cite{zhou2022conditional} which learns a dynamic prompt that is specifically designed to improve Base-to-New transfer. We also compare our CLIP-A-self against CLIP-A \cite{gao2021clip} due to the similarity in architecture and to show that the performance improvements are from making use of the visual sentences and not from the just adapting the text features.
\subsection{Training settings}
Our implementation is based on CoOp's and CLIP-A's code. \footnote{\url{https://github.com/KaiyangZhou/CoOp}, \url{https://github.com/gaopengcuhk/CLIP-Adapter}}
 We make all our comparisons on VIT CLIP backbone i.e., VIT-B/16. We take the results for CoOp and CoCoOp for all datasets (except CUB) from their respective papers, while we make use of practices from the respective papers like context length set to 4 and context initialization to ``a photo of" to ensure the best results on the CUB dataset. For CLIP-A, we re-run all experiments on VIT-B/16 backbone as they were not reported in the paper. For all adapter models including ours, we only tune the residual ratio \(\beta\) hyper-parameter. For CLIP-A, we use the version where the MLP is applied on top of the visual encoder as it performed the best \cite{gao2021clip}. We make use of May version of GPT-4 for obtaining the auxiliary dataset.

\subsection{GPT generated visual sentences improve 0-shot transfer.}

We compare the performance of CLIP-GPT prompt ensemble with the default prompts of CLIP in Table \ref{table_main_results_zs}. GPT-generated prompt ensemble improves upon the performance of CLIP 0-shot by ~2\(\%\) on average over 12 datasets. The improvement over CLIP-ZS is significant; over 5\(\%\) for specialized fine-grained datasets like CUB, SUN397, EuroSAT, and DTD and over 2\(\%\) for oxford-flowers and oxford-pets. This shows that CLIP does not recognize several of the classnames in these datasets and describing the class in the form of visually descriptive sentences results in better classifiers from the text-encoder and better classification accuracy. It is also worth noting that only including the visually descriptive sentences in the prompts can help improve the performance of general datasets like Imagenet (over 4\(\%\)) and Caltech-101 (over 1\(\%\)) too. For all other datasets, the transfer performance matches that of CLIP, with the exception being the action recognition dataset UCF-101. We inspected the sentences generated for UCF-101 and notice that several of the sentences generated by GPT involves temporal information instead of visual descriptions and we believe this could be the reason for the drop in accuracy. However, we notice in Section \ref{sec:attn_wts} that the self-attention module of the few-shot adapter learns to emphasize the visual sentences out of the generated sentences which might explain the improvement in the performance of few-shot adapters in the new setting in Section \ref{sec:adapter_results}. We also compare against recent work \cite{menon2022visual} on their subset of 6 datasets for VIT-B/16 encoder in \ref{table_subset_results_zs}. We see that using the larger GPT-4 model over the GPT-3 model results in much higher improvements for specialized datasets like DTD (\( \sim 5\%\)) and EuroSAT (\(\sim 6\%\)). We compare the text used by \cite{menon2022visual} against our GPT4-generated VDT in the supplementary.

% \subsection

\subsection{GPT-Adapters improve few-shot transfer performance.}
\label{sec:adapter_results}

We compare the performance of our CLIP-A-self against CLIP, CoOp, and CoCoOp on the benchmark suite of 12 datasets in the Base-to-New setting in Table \ref{table_results_generalization}. Here we see that GPT-Adapters that make use of the VDT information outperform CoCoOp by \(3\%\) in the new setting while maintaining similar performance to that of CoOp in the base setting on the average accuracy over 12 datasets. This is impressive considering that CoCoOp makes use of a meta-network and forward pass through the text encoder making it computationally intensive to train. CoCoOp takes up to 5 hours to train on 16-shot ImageNet for VIT-B/16 encoder, in comparison,  our CLIP-A-self takes only 10 mins (on an RTX 3090 GPU). The Base-to-New generalization ability of our adapters is even more impressive for fine-grained, specialized datasets as evidenced by the gains over CoCoOp in Harmonic mean of base and new accuracy. For example, CLIP-A-self demonstrates gains in datasets like FGVCAircraft (~7.5\(\%\)), EuroSat (~7.4\(\%\)), DTD (~5.8\(\%\)), CUB (~4.3\(\%\)), Flowers102 (~4\(\%\)),  Stanford Cars (~2.4\(\%\)) and UCF-101 (~2.4\(\%\) ). This demonstrates that our adapters make use of semantic information in the form of visually descriptive sentences and fuse this with CLIP's 0-shot knowledge to build more generalizable classifiers that transfer well to unseen classes within the same dataset. It is also worth noting that even though the same set of VDT did not provide any improvements in 0-shot domain transfer for datasets like FGVC-Aircraft, Stanford-Cars, and UCF-101, our self-attention adapter was able to choose the most informative subset of VDT and produce few-shot classifiers that provide substantial few-shot transfer performance gains in comparison to CoCoOp. We show in Section \ref{sec:attn_wts} the sentences picked by the attention mechanism for these datasets to qualitatively verify this.

%% comparison between mlp and attn adapter 
%%% what to write here.. Artificial experiment where we add non-visual information and it shows how accuracy of mlp reduces while accuracy of attn stays the same.

\subsubsection{Attention weights Analysis}
\label{sec:attn_wts}
We note that even though CLIP-gpt ensembles were outperformed by CLIP default prompt on FGVC Aircraft, UCF-101, and Stanford Cars dataset, we see that CLIP-A-self outperforms CLIP-A and CoCoOp \cite{zhou2022conditional} on these datasets in the few-shot transfer setting. We believe that this is because, during few-shot training, the self-attention mechanism learns to select the most relevant visual sentences out of the set of visually descriptive text and helps produce generalizable classifiers. In Table 1 in supplementary, we show the top 3 and bottom 3 attributes picked by attention scores for each of these datasets and show that the sentences with the highest attention scores correspond to visually descriptive attributes in the set and vice versa for the lowest scored attributes. For example, for both Stanford Cars and FGVC it is interesting to see that the color scheme is one of the least used attributes as it's difficult to identify a car or a plane from its color or livery. For UCF-101, information like the force involved or temporal information like speed and range of motion of the action is unlikely to be encoded in the image and hence is not selected by the attention mechanism. Information regarding the subject and the object of the action, like the posture of the person, description of the object, and interaction between objects are visible in the images and hence weighted highly by the attention mechanism.

\subsection{Ablation over different GPT models}
In this section, we see if other GPT models like GPT-3.5 and open-source model, OpenAssistant \cite{kopf2023openassistant}, are as capable as GPT-4 in generating visually descriptive information. We explore this on the CUB dataset as it is fine-grained and specialized. The results are presented in Table \ref{table:gpt_models}. We find that the performance improves with larger models which are more capable of memorizing accurate class information with less hallucination \cite{YuKoLA:Models}. Even though we obtain decent performance with the open-source model OpenAssistant, the outputs were always inconsistent and noisy, resulting in a lot of clean-up effort in comparison to GPT-3.5 and GPT-4 where the outputs were in the form of concise sentences following a dictionary format.
It is worth noting that our few-shot adapter is capable of picking out the the best VDT information even from a noisy set, pushing the Base-to-New generalization performance of OpenAssistant, and GPT-3.5 close to that of GPT-4. 
% We provide results for all datasets in the benchmark using GPT-3.5 in the supplementary material.

% \subsection{Ablation over the number of heads in self-attention}

% \subsection{Lower shots results}

%% run imagenet for coop and co-co-op at 1 shot and 5-shot and 10-shot and show how our method is better? %% not so impressive. so leave it out, add the plots in supplementary. write about it in results

%------------------------------------------------------------------------

%-------------------------------------------------------------------------

%-------------------------------------------------------------------------

%------------------------------------------------------------------------
\section{Conclusion}
In this work, we show that using visually descriptive textual (VDT) information can improve the 0-shot domain transfer performance of CLIP over non-visual information and the default prompts. We demonstrate GPT-4 to be an accurate and flexible source of VDT information by improving the 0-shot domain transfer performances on a suite of 12 benchmark datasets. Our few-shot adapter CLIP-A-self learns to pick the best VDT information from the GPT generated set and improve the few-shot domain transfer in the Base-to-New setting even when the quality of the generated text deteriorates. We release all prompts and VDT information for all 12 datasets to promote further research in the fertile research direction of using LLMs for learning multi-modal adapters for foundation models.

{\small
\bibliographystyle{ieee_fullname}
\bibliography{egbib}
}

%%%%%%%%%% Merge with supplemental materials %%%%%%%%%%
\pagebreak
% \newpage
\clearpage
% \widetext
\begin{center}
\textbf{\large Supplementary Materials: Enhancing CLIP with GPT-4: Harnessing Visual Descriptions as Prompts}
\end{center}

\setcounter{equation}{0}
\setcounter{section}{0}
\setcounter{figure}{0}
\setcounter{table}{0}
\setcounter{page}{1}
\makeatletter
\renewcommand{\theequation}{S\arabic{equation}}
\renewcommand{\thefigure}{\arabic{figure}}

\begin{table*}[]
\resizebox{\textwidth}{!}{%
\begin{NiceTabular}{l|l|l}
\toprule
\textbf{Dataset} & \textbf{Top 3 attributes selected}                                                                                        & \textbf{Bottom 3 attributes selected}                                                    \\
\midrule
\rule{0pt}{2ex}
FGVC    & \makecell[tl]{Unique visual identifier, \\ presence of canards, \\ tail type.}                                     & \makecell[tl]{Color scheme,\\ model number, \\ commercial or cargo.}             \\
\rule{0pt}{3ex}    
Cars    & \makecell[tl]{Body shape, \\ fender description, \\ spoiler description,}                                          & \makecell[tl]{Interior description, \\ brand logo description, \\ color scheme}   \\
\rule{0pt}{3ex}    
UCF-101 & \makecell[tl]{Equipment used, \\ 
Posture of person, \\ Interaction info.} &  \makecell[tl]{Body muscles used, \\force involved, \\ speed of motion}  \\
\rule{0pt}{3ex}    
Oxford-Flowers & \makecell[tl]{Shape of the flower, \\Color, shape and number of petals,\\ Texture and description of veins in leaves} & \makecell[tl]{Stem color,\\ Color of leaves,\\ Description of sepals}
\\
\rule{0pt}{3ex}    
CUB & \makecell[tl]{Wings color and shape, \\ Head color and shape,\\ Beak color 
 and shape} & \makecell[tl]{Color and description of legs, \\Underparts color,\\ Tail shape and color}
% \bottomrule
\end{NiceTabular}
}%
\caption{The top 3 and bottom 3 attributes selected by the attention mechanism in GPT-A-self for 3 different datasets. For UCF101, We see that attention learns to pick visually descriptive sentences like posture and description of objects over temporal information like speed of motion and force applied.}
\label{table:attn_wts}
\end{table*}

\section{Attention weights visualized}
We visualize the attention weights learned by the CLIP-A-self for datasets Stanford Cars, UCF101, FGVC Aircraft, Oxford Flowers and CUB in Table \ref{table:attn_wts}. 
We notice that the self-attention mechanism in CLIP-A-self  assigns more weight to visually descriptive sentences that are most relevant for discriminating 
between the classes of the dataset under consideration. For instance, we see that for discriminating images of birds species (CUB dataset) and flower species (Oxford Flowers)
sentences describing the color of the head and wings of birds and petals of the flowers are important but for identifying different car or aircraft models sentences
describing the color or livery is one of the least important. We also see that if the information being described by the VDT sentence is not clearly visible in the image,
the attention weight assigned to it by CLIP-A-self is low. For instance, in CUB dataset, the the undersides of birds or the sepals in Oxford Flowers dataset are often not visible in the images, hence 
the VDT sentence correponding to this is is in the bottom 3 attributes picked by the learnt attention weights. It's also worth noting that, some of the VDT sentences 
do not have much variation between different classes and hence are not useful in dsicrimination between the classes of the dataset. For instance, in Oxford-flowers,
the color of the leaves, the color of the stem are often green for most flowers in the dataset, which maybe why low attention score was learnt for this attribute.

\section{Prompts for GPT-4}

Throughout our experiments, we use a two-step prompting strategy in which we first ask the LLM to generate a list of attributes which will aid in visually distinguishing between the different classes in a particular dataset. The second prompt asks the LLM to create a description using the attributes provided by the first prompt and specifies the expected output format. We request a python dictionary as output with a list of sentences, each corresponding to one attribute. The output structure is simple to use (downstream), preserves attribute-level detail and encourages attribute richness. 

Example of first prompt for the FGVC Aircraft dataset:
\begin{quote}\textit{
I am creating class attributes for a zero-shot image recognition algorithm to classify different images of a diverse set of aircrafts. The attributes are part of side information about the classes. List 20 attributes that can form part of a description of the class that will aid in distinguishing between the following list of classes visually:
[list of classes in dataset e.g. 'Boeing 707-320', 'Boeing 727-200',..., 'Yakovlev Yak-42']}
\end{quote}

GPT-4 response:

\begin{quote}
    \textit{
    Manufacturer: The company that produces the aircraft (e.g., Boeing, Airbus). 
Model Number: Differentiates different aircraft within the same manufacturer. 
Aircraft Function: The primary role or use of the aircraft, such as commercial (passenger), cargo (freight transport), or military (fighter, bomber, etc). 
Engine Count: The number of engines on the aircraft. 
Engine Position: The location of the engines, either under the wings or at the rear of the fuselage. 
Wing Position: The position of the wings: high-wing, mid-wing, or low-wing design. 
Tail Type: The shape and type of the tail, such as traditional, T-tail, V-tail, etc. 
Body Length: The length of the aircraft's fuselage. 
Body Shape: The overall shape of the fuselage, like narrow, wide, double-decker, etc. 
Wing Shape: The shape of the wings, either swept or straight. 
Nose Shape: The shape of the nose or front of the aircraft. 
Landing Gear Type: The type of landing gear, like tricycle or taildragger. 
Cockpit Window Shape: The shape and number of cockpit windows. 
Color Scheme: Typical color schemes or liveries associated with the aircraft type. 
Number of Decks: Either single-deck or double-deck (like the Airbus A380). 
Existence of Winglets: Whether or not the aircraft has winglets at the end of the wings. 
Existence of Canards: Whether or not the aircraft has canards, small wing-like surfaces located at the front. 
Propeller or Jet: Indicates if the aircraft is propeller-driven or jet-powered. 
Number of Seats: An estimate of the seating capacity of the aircraft. 
Range: The maximum distance the aircraft can travel without refueling. 
Unique Visual Identifier: Specific structures or features that are unique to a model or series of aircraft. 
Aircraft Type: Differentiates between commercial, cargo, or military aircraft.
    }
\end{quote}

\begin{quote}
    
\end{quote}

Example of the second prompt for the FGVC Aircraft dataset which uses GPT-4's response for the first prompt as part of the prompt:

\begin{quote}\textit{
Describe the following aircraft by adding one sentence about each attribute for the following aircraft: Airbus A340-200. Return the answer as a python dictionary with the aircraft variant as the key (i.e. remove the manufacturer) and the value is a list of sentences. Rewrite the attribute as a full sentence. Do not include the attributes as keys. Attributes: 
Manufacturer: The company that produces the aircraft (e.g., Boeing, Airbus). 
Model Number: Differentiates different aircraft within the same manufacturer. 
Aircraft Function: The primary role or use of the aircraft, such as commercial (passenger), cargo (freight transport), or military (fighter, bomber, etc). 
Engine Count: The number of engines on the aircraft. 
Engine Position: The location of the engines, either under the wings or at the rear of the fuselage. 
Wing Position: The position of the wings: high-wing, mid-wing, or low-wing design. 
Tail Type: The shape and type of the tail, such as traditional, T-tail, V-tail, etc. 
Body Length: The length of the aircraft's fuselage. 
Body Shape: The overall shape of the fuselage, like narrow, wide, double-decker, etc. 
Wing Shape: The shape of the wings, either swept or straight. 
Nose Shape: The shape of the nose or front of the aircraft. 
Landing Gear Type: The type of landing gear, like tricycle or taildragger. 
Cockpit Window Shape: The shape and number of cockpit windows. 
Color Scheme: Typical color schemes or liveries associated with the aircraft type. 
Number of Decks: Either single-deck or double-deck (like the Airbus A380). 
Existence of Winglets: Whether or not the aircraft has winglets at the end of the wings. 
Existence of Canards: Whether or not the aircraft has canards, small wing-like surfaces located at the front. 
Propeller or Jet: Indicates if the aircraft is propeller-driven or jet-powered. 
Number of Seats: An estimate of the seating capacity of the aircraft. 
Range: The maximum distance the aircraft can travel without refueling. 
Unique Visual Identifier: Specific structures or features that are unique to a model or series of aircraft. 
Aircraft Type: Differentiates between commercial, cargo, or military aircraft.}
\end{quote}

The response of the second prompt constitutes the VDT information we utilise as side-information for Airbus A340-200 as an example:

\begin{quote}
    \textit{
    "A340-200": [
"The Airbus A340-200 is produced by Airbus, a renowned aircraft manufacturer.",
"It differentiates itself from other aircraft within the Airbus family through its unique model number: A340-200.",
"This aircraft primarily serves a commercial role, typically used for passenger transport.",
"The Airbus A340-200 is equipped with four engines.",
"These engines are situated under the aircraft's wings.",
"The aircraft features a low-wing design, with wings positioned at the bottom of the fuselage.",
"It has a traditional tail configuration, common to many large commercial aircraft.",
"The A340-200 has a lengthy fuselage, extending to about 59.4 meters.",
"The body of the Airbus A340-200 is wide-bodied, facilitating a larger passenger capacity.",
"Its wings are swept back, a design aspect that improves fuel efficiency and performance at high speeds.",
"The aircraft features a rounded nose shape, contributing to its aerodynamic design.",
"The Airbus A340-200 uses a tricycle type landing gear, supporting stability during takeoffs and landings.",
"Its cockpit windows are angular and include six panels, giving pilots a comprehensive view of their surroundings.",
"Color schemes vary by airline, but the Airbus corporate livery features a predominantly white body with blue and teal accents.",
"This model is a single-deck aircraft, focusing on width rather than height for passenger capacity.",
"The A340-200 does not have winglets, differing from some newer Airbus models.",
"There are no canards present on the Airbus A340-200, instead, it employs a more traditional aircraft design.",
"As a jet-powered aircraft, the A340-200 uses high-speed jet engines for propulsion.",
"The A340-200 typically accommodates around 260 passengers, though the exact number can vary with the configuration.",
"With a range of approximately 7,800 nautical miles, the Airbus A340-200 can cover considerable distances without refueling.",
"The aircraft's four-engine configuration and lengthy, wide-bodied design are unique visual identifiers of the A340-200 model.",
"Classified as a commercial aircraft, the Airbus A340-200 is primarily used for passenger transportation."
]
    }
\end{quote}

GPT-4 generally adheres to the python dictionary output requirement in the \textit{User} prompt, but tends to return additional explanations, motivations or clarifications. To encourage the LLM to only return a Python dictionary as requested, we add the following \textit{System} prompt:
\begin{quote}\textit{
You are ChatGPT, a large language model trained by OpenAI. Return only the python dictionary, with no explanation.  }
\end{quote}

Conversely, OpenAssistant's \cite{kopf2023openassistant} output requires manual cleaning and reformatting to get into Python dictionary format. GPT-3.5 performed slightly worse than GPT-4 in terms of adherence to the prompt, as it did not consistently return only a dictionary. In such cases, we simply called the API again. After repeated incorrect format responses, we manually cleaned those cases.

We primarily utilized GPT-4 via the ChatGPT Plus subscription plan at a cost of \$20 since the GPT-4 API was not generally available during most of our experimentation phase.
The GPT-4 API cost to create the VDT information for the SUN397 dataset was \$14.90, as opposed to \$1.94 using the GPT-3.5 API.

% Open Assistant 

\section{Comparing our VDT with GPT3}
In Table 2, we compare the VDT generated by GPT-4 using our prompting technique with that of \cite{menon2022visual} who used GPT-3 to obtain visual descriptors for 
different classes of the dataset. Here we notice that, including a prompt step asking the GPT-4 for visual attributes necessary for classifying between images of the 
classes result in a fixed number of sentences per class, a fixed order guaranteeing that every class is accompanied by as much visual information as possible. By using GPT-4
we also get much richer and more accurate visual descriptions. For example, for the class industrial, our descriptions provide inforrmation about density of buildings, 
shadow in the image, road accessibility and layout while the description used by \cite{menon2022visual} is only `evidence of human activity'. A similar phenomenon
can be observed for DTD dataset. This explains the jump in performance for specialized datasets like DTD and Eurosat over DCLIP.

\section{Generalizability at lower shots}
In Figure 1, we compare the harmonic mean of Base and New accuracies of CLIP-A-self with that of CLIP-A over number of shots = 1, 5, 10, 16.  
Our CLIP-A-self demonstrates   performance improvements at lower shots, outperforming CLIP-A on average by over \(1.5\%/\) for the 1-shot case and over   \(2.5\%/\) for the 5-shot case. Our adapter shows higher improvements over CLIP-A in the higher shot  scenario because of the number of parameters and the inherent difficulty in identifying the VDT sentences that are discriminative for the current classes in the low shot scenario. For instance, identifying the class from a single image is often difficult because of co-occuring objects, environment, background etc which can be resolved if we have more exmaple images from the same class. The largest improvements are for specialized  and fine-grained datasets like Stanford-Cars, EuroSat Oxford Flowers, DTD and CUB. Oxford-pets and Food-101 results do not improve much because these datasets are relatively easy and already show good performance with default CLIP. 

%% improve this. maybe add more.

% \section{Ablation over the percentage of VDT}
%% can add this if we want. Synthetic experiment where we mix VDT and non VDT, plot the zs and adapter hmean acccuracy vs % of visual information for a few datasets.

\begin{table*}[]
    \centering
    \caption{Comparing our VDT with that of descriptors from \cite{menon2022visual} for 2 random classes of datasets DTD and Eurosat}
    \resizebox{\textwidth}{!}{%
        \begin{tabular}{|l|c|c|}
            \hline
            & \textbf{Ours} & \textbf{DCLIP\cite{menon2022visual}} \\
            \hline
            \multirow{8}{*}{\textbf{Stratified (DTD)}} & 'The surface feels moderately smooth, with slight roughness due to the layered structure.' & 'a series of layers' \\
            & 'There is no distinct pattern, but the layers create a natural, linear visual effect.' & 'each layer is of a different material' \\
            & 'The structure is characterized by multiple layers stacked upon each other.' & 'the layers are parallel to each other' \\
            & 'The texture has a two-dimensional feel, with the layers adding a sense of depth.' & 'the layers may be of different thicknesses' \\
            & 'The density varies, with some layers appearing closely packed while others are more sparse.' & 'the layers may be of different colors' \\
            & 'The regularity of the texture is defined by the consistent layering.' & 'the layers may have different textures' \\
            & 'The texture is opaque, with no transparency between the layers.' & \\
            & 'There are no significant surface defects, but minor irregularities may occur between layers.' & \\
            \hline
            \multirow{8}{*}{\textbf{Lined (DTD)}} & 'The texture feels moderately smooth to the touch, not too rough nor too sleek.' & 'a series of parallel lines' \\
            & 'It exhibits a lined pattern, reminiscent of ruled notebook paper.' & 'can be straight or curved' \\
            & 'The structure of the texture is stratified, with lines arranged one after the other.' & 'may be of different colors' \\
            & 'The texture has a two-dimensional quality, with no noticeable depth or relief.' & 'may be of different widths' \\
            & 'The lines are densely packed, leaving little space between them.' & 'may be of different thicknesses' \\
            & 'The texture displays a high degree of regularity, with the lines evenly spaced and parallel.' & \\
            & 'The texture is opaque, with no transparency or translucency.' & \\
            & 'There are no noticeable surface defects, the lines are clean and uninterrupted.' & \\
            \hline
            \multirow{11}{*}{\textbf{Industrial (Eurosat)}} & 'Industrial buildings have texture that is smooth, regular.' & 'evidence of human activity' \\
            & 'Industrial buildings have shape that is rectangular, irregular.' & \\
            & 'Industrial buildings have size (relative) that is large.' & \\
            & 'Industrial buildings have pattern that is regular, dense.' & \\
            & 'Industrial buildings have spectral reflectance that is high in visible spectrum.' & \\
            & 'Industrial buildings have a shadow that is present (due to high-rise buildings).' & \\
            & 'Industrial buildings have adjacent land features that is commercial, residential, roads.' & \\
            & 'Industrial buildings have change over time that is stable.' & \\
            & 'Industrial buildings have density that is high.' & \\
            & 'Industrial buildings have proximity to water bodies that is variable.' & \\
            & 'Industrial buildings have road accessibility that is high.' & \\
            \hline
            \multirow{11}{*}{\textbf{Forest (Eurosat)}} & 'Forest has texture that is rough.' & 'a large area of trees' \\
            & 'Forest has shape that is irregular.' & 'green leaves' \\
            & 'Forest has size (relative) that is large.' & \\
            & 'Forest has pattern that is no pattern.' & \\
            & 'Forest has spectral reflectance that is high in near-infrared.' & \\
            & 'Forest has shadow that is present (due to trees).' & \\
            & 'Forest has adjacent land features that is land, mountains, rivers.' & \\
            & 'Forest has change over time that is mostly stable.' & \\
            & 'Forest has density that is high.' & \\
            & 'Forest has proximity to water bodies that is variable.' & \\
            & 'Forest has road accessibility that is low.' & \\
            \hline
        \end{tabular}%
    }%
\end{table*}

\begin{figure*}[t]
    \centering
    \begin{minipage}[t]{0.325\linewidth}
    \centering
    \includegraphics[width=1.7in]{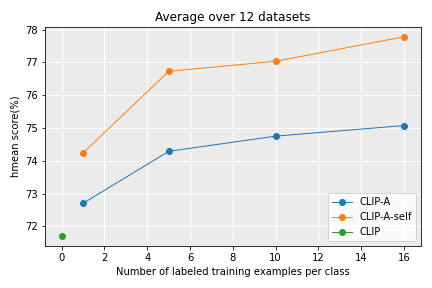}
    \end{minipage}
    \begin{minipage}[t]{0.325\linewidth}
    \centering
    \includegraphics[width=1.7in]{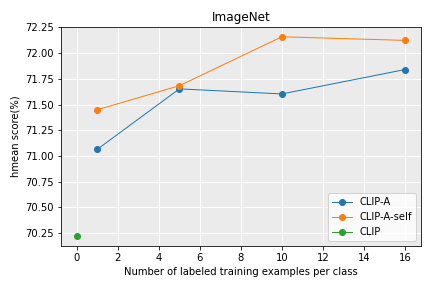}
    \end{minipage}
    \begin{minipage}[t]{0.325\linewidth}
    \centering
    \includegraphics[width=1.7in]{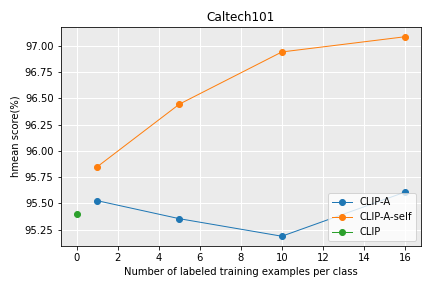}
    \end{minipage}
    
    \hspace{0.2in}
    
    \begin{minipage}[t]{0.325\linewidth}
    \centering
    \includegraphics[width=1.7in]{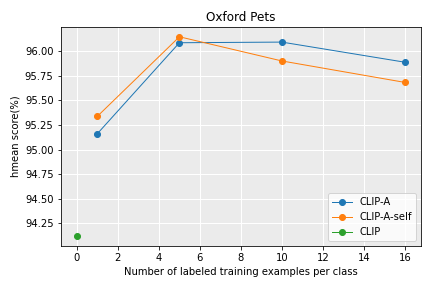}
    \end{minipage}
    \begin{minipage}[t]{0.325\linewidth}
    \centering
    \includegraphics[width=1.7in]{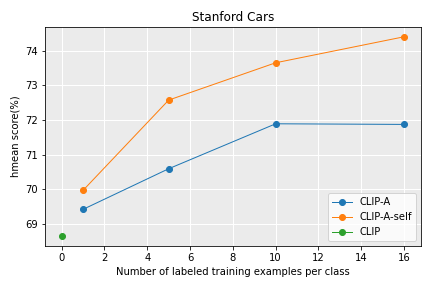}
    \end{minipage}
    \begin{minipage}[t]{0.325\linewidth}
    \centering
    \includegraphics[width=1.7in]{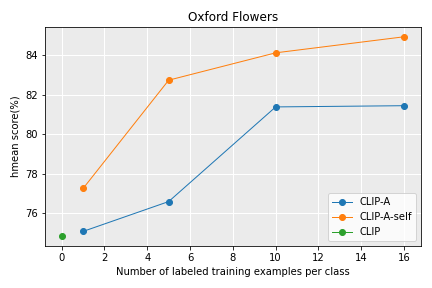}
    \end{minipage}
    
    \hspace{0.2in}
    
    \begin{minipage}[t]{0.325\linewidth}
    \centering
    \includegraphics[width=1.7in]{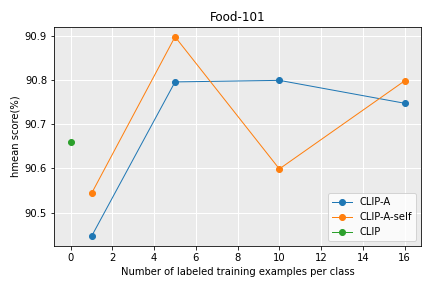}
    \end{minipage}
    \begin{minipage}[t]{0.325\linewidth}
    \centering
    \includegraphics[width=1.7in]{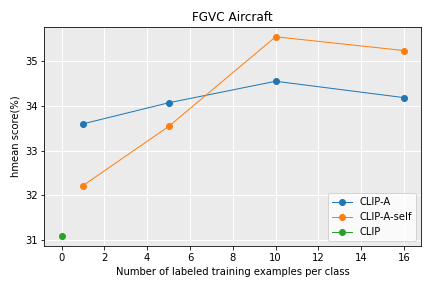}
    \end{minipage}
    \begin{minipage}[t]{0.325\linewidth}
    \centering
    \includegraphics[width=1.7in]{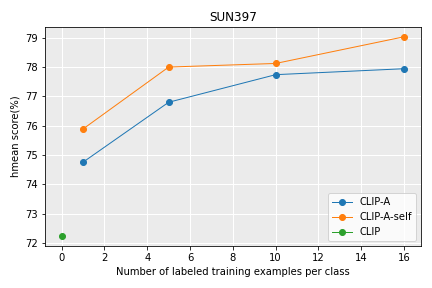}
    \end{minipage}
    
    \hspace{0.2in}
    
    \begin{minipage}[t]{0.325\linewidth}
    \centering
    \includegraphics[width=1.7in]{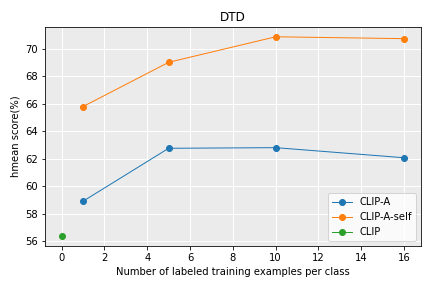}
    \end{minipage}
    \begin{minipage}[t]{0.325\linewidth}
    \centering
    \includegraphics[width=1.7in]{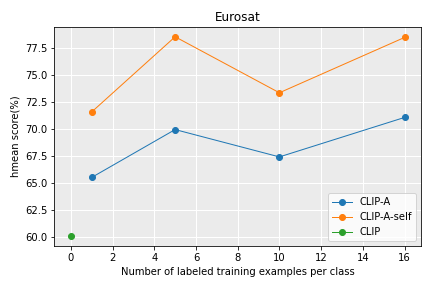}
    \end{minipage}
    \begin{minipage}[t]{0.325\linewidth}
    \centering
    \includegraphics[width=1.7in]{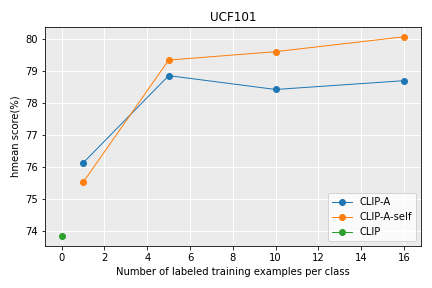}
    \end{minipage}

    \begin{minipage}[t]{0.325\linewidth}
    \centering
    \includegraphics[width=1.7in]{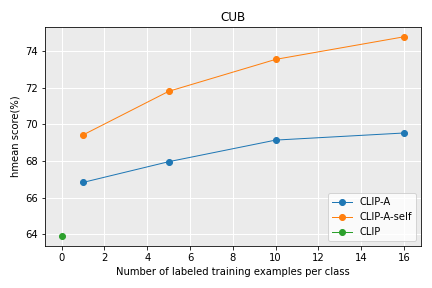}
    \end{minipage}
    \centering
    \caption{Main results of Base-to-New few shot learning on 12 datasets. CLIP-A-self consistently shows better performance over CLIP-A over different training shots, demonstrating the importance of Visually descriptive text in improving the generalizability of few-shot classifiers for CLIP. }
    \label{fig:main_results}
    \vspace*{-12pt}
\end{figure*}

\end{document}